%% file: icme.tex
\documentclass[conference]{IEEEtran}
\IEEEoverridecommandlockouts
\usepackage{cite}
\usepackage{amsmath,amssymb,amsfonts}
\usepackage{algorithmic}
\usepackage{graphicx}
\usepackage{textcomp}
\usepackage{xcolor}
\usepackage{booktabs}
\usepackage{colortbl}

\definecolor{brown}{HTML}{E6C8B3}
\definecolor{yellow}{HTML}{FBEBDD}

\def\BibTeX{{\rm B\kern-.05em{\sc i\kern-.025em b}\kern-.08em
    T\kern-.1667em\lower.7ex\hbox{E}\kern-.125emX}}
\definecolor{weblink}{HTML}{F60286}
\usepackage[pagebackref,breaklinks,colorlinks=true, citecolor=black,urlcolor=weblink]{hyperref}
\begin{document}

\title{Expansive Supervision for Neural Radiance Fields}

\author{
Weixiang Zhang\textsuperscript{\rm 1},
Wei Yao\textsuperscript{\rm 1},
Shuzhao Xie\textsuperscript{\rm 1}, 
Shijia Ge\textsuperscript{\rm 1},
Chen Tang\textsuperscript{\rm 2}, 
Zhi Wang\textsuperscript{\rm 1*} \\
\textsuperscript{\rm 1}Tsinghua Shenzhen International Graduate School, Tsinghua University\\
\textsuperscript{\rm 2}The Chinese University of Hong Kong \\

\small \{zhang-wx22, w-yao22, gsj23, xsz24, wmz22\}@mails.tsinghua.edu.cn, chentang@link.cuhk.edu.hk, wangzhi@sz.tsinghua.edu.cn
}

\maketitle{
\renewcommand{\thefootnote}{\fnsymbol{footnote}}
\footnotetext[1]{Corresponding author.}
}

\input{content/0-abstract}


\input{content/1-introduction}

\input{content/3-methods}

\input{content/4-experiments}

\input{content/5-conclusion}

\section*{Acknowledgment}

National Key Research and Development Project of China (Grant No. 2023YFF0905502), National Natural Science Foundation of China(Grant No. 92467204 and 62472249), and Shenzhen Science and Technology Program (Grant No. JCYJ20220818101014030 and KJZD20240903102300001).

\bibliographystyle{IEEEbib}
\bibliography{icme2025references}

\end{document}

%% file: content/0-abstract.tex
\begin{abstract}
Neural Radiance Field (NeRF) has achieved remarkable success in creating immersive media representations through its exceptional reconstruction capabilities. 
However, the computational demands of dense forward passes and volume rendering during training continue to challenge its real-world applications.
In this paper, we introduce Expansive Supervision to reduce time and memory costs during NeRF training from the perspective of partial ray selection for supervision. 
Specifically, we observe that training errors exhibit a long-tail distribution correlated with image content. Based on this observation, our method selectively renders a small but crucial subset of pixels and expands their values to estimate errors across the entire area for each iteration. Compared to conventional supervision, our approach effectively bypasses redundant rendering processes, resulting in substantial reductions in both time and memory consumption.
Experimental results demonstrate that integrating Expansive Supervision within existing state-of-the-art acceleration frameworks achieves 52\% memory savings and 16\% time savings while maintaining comparable visual quality.
Our code is available at \textit{\href{https://github.com/zwx-open/Expansive-Supervision}{this link}}.

\end{abstract}

%% file: content/1-introduction.tex
\section{Introduction}

\begin{figure}[!t]
    \centerline{\includegraphics[width=\columnwidth]{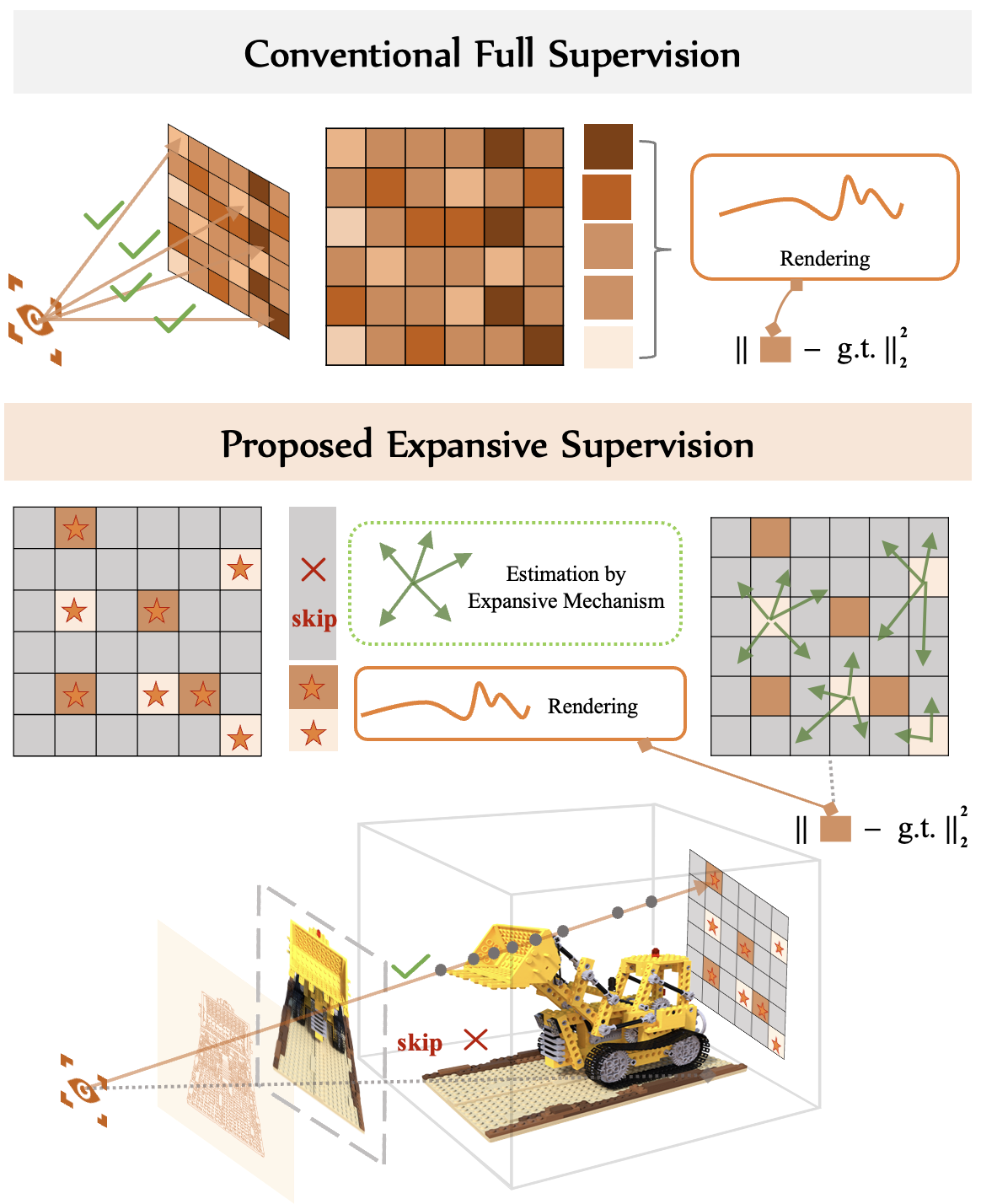}}
    \caption{\textbf{Overview of proposed method.} Our approach adopts an expansive supervision technique to selectively render a subset of crucial pixels to estimate the error by expansive mechanism.
    Unlike conventional full supervision, which blindly renders all pixels, our method intelligently avoids redundant rendering processes, leading to significant reductions in training time and memory consumption.
    }
    \label{fig:observation}
\end{figure}

Neural Radiance Field (NeRF)~\cite{nerf} has emerged as a promising approach for representing 3D media content in the field of photorealistic novel view synthesis.
NeRF employs a neural network $F(\Theta)$ to implicitly encode scenes by mapping position ${\mathbf{x}}=(x, y, z)$ and direction $\mathbf{d}=(\theta, \varphi)$ to view-dependent color ${\mathbf{c}}=(r, g, b)$ and view-independent volumetric density $\tau$, $\text{i.e.}, \ F(\Theta):(\mathbf{x}, \mathbf{d}) \rightarrow(\mathbf{c}, \sigma)$, where $\Theta$ is parameters of the neural network.
Through its powerful implicit neural scene representation, NeRF synthesizes and infers target pixels via volume rendering using sampled query pairs~(color $\mathbf{c}$ and density $\sigma$) along rays, significantly surpassing traditional multi-view stereo methods in view synthesis and 3D construction.

Despite the impressive performance of NeRF, its training speed remains a significant limitation. In the original NeRF architecture, rendering each pixel requires sampling $N$ points to compute the color $\mathbf{c}$ and density $\sigma$. For a scene with dimensions $(h,w)$, this process necessitates $h \cdot w \cdot N$ neural network forward passes, potentially exceeding $10^{8}$ computations for rendering a single 1080p resolution view. This computational burden substantially prolongs the training duration and impedes the practical generation of neural 3D media content.
To address these computational challenges, various acceleration methods have been proposed, including decomposition schemes~\cite{kilo}, distillation to light fields~\cite{mobile-r2l}, and incorporation of explicit representations~\cite{tensorf, ingp}.
Among these, the most effective approaches leverage explicit representations to cache view-independent features, trading memory consumption for reduced training time.
These efficient explicit representations include voxel grids~\cite{dvgo}, hash tables~\cite{ingp}, and low-rank tensors~\cite{tensorf}. By incorporating such explicit structures, training time for a scene can be reduced significantly, from approximately 10 hours to 30 minutes.

\begin{figure*}[!t]
    \centerline{\includegraphics[width=\textwidth ]{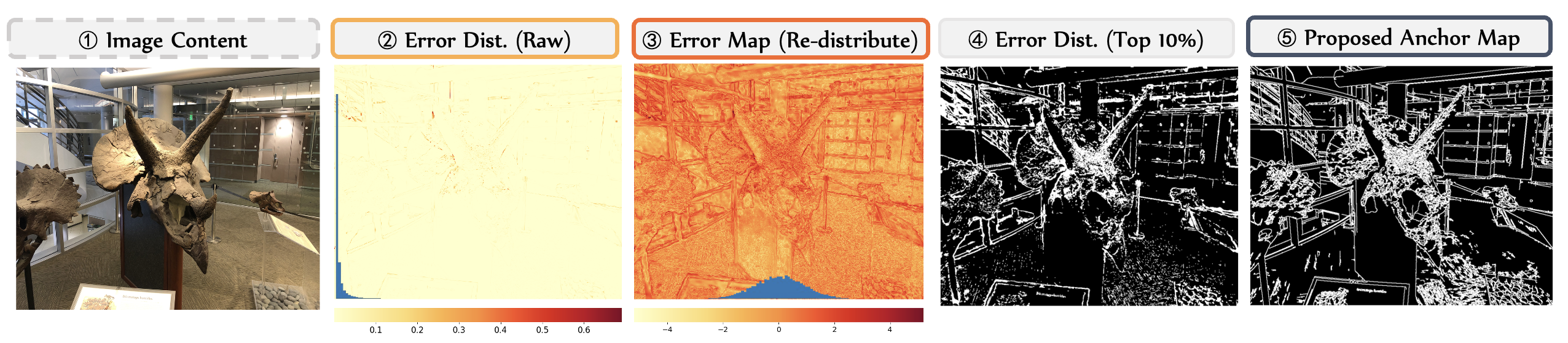}}
    \caption{
    \textbf{Motivational observation.}
    (\#2) The blue histogram shows the error distribution after 1000 iterations, revealing a pronounced long-tail characteristic.
    (\#3) To enhance the visibility of error map, we transformed the data into a normal distribution, revealing correlations between redistributed errors and image content.
    (\#4) The top 10\% of errors identified during training are visualized, corresponding to regions with high-frequency details in the image content. 
    (\#5) The top 10\% error map generated by our expansive supervision exhibits a high correlation with the actual error distribution.
    }
    \label{fig:observation}
\end{figure*} 

Differing from previous methods, we propose Expansive Supervision, an acceleration method for NeRF training based on ray selection and partial supervision, which can be integrated with existing acceleration frameworks to further reduce time and memory consumption. Our approach is motivated by the observation that training errors exhibit a long-tail distribution strongly correlated with image content: regions with higher frequencies display larger errors, while smoother areas show smaller errors.
This observation leads us to selectively render a small but crucial subset of pixels ${R}^{\prime} \subset {R}$ guided by image content prior during training, and expand the error of these precisely rendered pixels to estimate the loss for the entire area in each iteration. Specifically, the selected pixel set $R^{\prime}$ comprises two areas: the anchor area $A$ and the source area $S$. The anchor area is determined by a meticulously designed frequency prior extractor to capture prominent error patterns, while the source area is sampled to expand its values to the remaining regions to maintain reconstruction quality. The final error estimate $\hat{L}$ is synthesized from the selected pixels ${R}^{\prime} =\{A \cup S \}$ through the expansive mechanism, and the model is updated via $\Theta:=\Theta-\eta \nabla \hat{L}({R}^{\prime})$, where $\eta$ denotes the learning rate.
By avoiding costly yet marginal renderings, our method theoretically achieves computational savings proportion to $(1-\frac{|R^{\prime}|}{|R|})$. Experimental results demonstrate that by rendering only 50\% of pixels to supervise the entire model, we achieve 0.52$\times$ memory and 0.16$\times$ time savings.

Our method shares similar mechanisms with recent sampling-based neural field optimization approaches, such as EGRA~\cite{egra} and Soft Mining~\cite{soft}, which resample rays in each batch based on image edge areas and Langevin Monte Carlo (LMC) sampling, respectively. While these methods achieve better reconstruction quality than uniform sampling with the same number of iterations, our method distinctively reduces both memory consumption and training time simultaneously. Furthermore, under identical conditions, our method demonstrates superior reconstruction quality compared to these sampling-based methods.

Extensive experiments validate the effectiveness of our proposed method. Compared to conventional full supervision, our approach achieves substantial reductions in both time and memory usage while maintaining comparable rendering quality. Notably, our method seamlessly integrates with existing explicit caching acceleration frameworks without requiring custom modifications.
Our main contributions are summarized as follows:
\textbf{(1)}~We observe a strong correlation between error distribution and image content, motivating the development of partial supervision methods to accelerate NeRF training within existing acceleration frameworks.
\textbf{(2)}~We propose Expansive Supervision, which selectively renders a small yet crucial subset of pixels for acceleration. This method significantly reduces training time and memory consumption by eliminating redundant rendering operations.
\textbf{(3)}~We present comprehensive experimental results validating the effectiveness of our method and detailed analysis of quality-cost trade-offs.

%% file: content/3-methods.tex
\section{Methods}
\subsection{Preliminary and Formulation}
\label{sec:formulation}
\label{sec:formulation}
Neural Radiance Field learns a function $\ F(\Theta):(\mathbf{x}, \mathbf{d}) \rightarrow(\mathbf{c}, \sigma)$ using a multilayer perceptron~(MLP), where ${\mathbf{x}} \in   \mathbb{R}^3, \mathbf{d} \in   \mathbb{R}^2$ represent the position and view direction of a point, while $\mathbf{c} \in \mathbb{R}^3$ and $\sigma \in \mathbb{R}$ represent the emitted color and density, respectively.
Volume rendering allows computing the expected color $\mathbf{C}(\mathbf{r})$ in a novel view as:
\begin{equation}
    \label{eq:render}
    \mathbf{C}(\mathbf{r}) = \int_0^t \mathcal{T}(t ; \mathbf{r}) \cdot \tau(\mathbf{r}(t)) \cdot \mathbf{c}(\mathbf{r}(t), \mathbf{d}) dt,
\end{equation}
where $\mathcal{T}(t ; \mathbf{r}) = \exp \left(-\int_0^t \tau(\mathbf{r}(s)) ds\right)$ represents the accumulated transmittance along the ray $\mathbf{r}(t) = \mathbf{o} + t \mathbf{d}$. 

In practice, numerical estimation of the rendering integral involves sampling $N$ points from partitioned bins along the ray, allowing the estimation of $\mathbf{C}(\mathbf{r})$ as:
\begin{equation}
    \label{eq:render2}
    \hat{\mathbf{C}}(\mathbf{r}) = \sum_{i=1}^N \mathcal{T}_i \left(1-\exp \left(-\tau_i \delta_i\right)\right) \mathbf{c}_i,
\end{equation} 
where $\mathcal{T}_i = \exp \left(-\sum_{j=1}^{i-1} \tau_j \delta_j\right)$, and $\delta_i$ represents the distance between adjacent sampled points.

The training of radiance fields is computationally intensive due to the large number of required neural network forward passes, which amounts to $N \times b$ for each iteration, where $N$ and $b$ denotes the number of sampled points along the ray and batch size.
Our method aims to reduce the computational resources by selecting a subset of pixels $R^{\prime}\in R$ to supervise the model, bypassing redundant renderings.
We utilize image context $\mathcal{I}$ to estimate the total error $L(R)$ with the expansive mechanism based on estimated error $\hat{L}(R^{\prime})$ from partial significant pixels $R^{\prime}$. 
This mechanism allows for the conservation of both time and memory, resulting in a savings of $(1-\frac{|R^{\prime}|}{|R|})v\times$ computational resources, where $v \in (0,1)$ represents the ratio of resources used by volume rendering in the total training process.

 \begin{figure}[t]
    \centerline{\includegraphics[width=\columnwidth ]{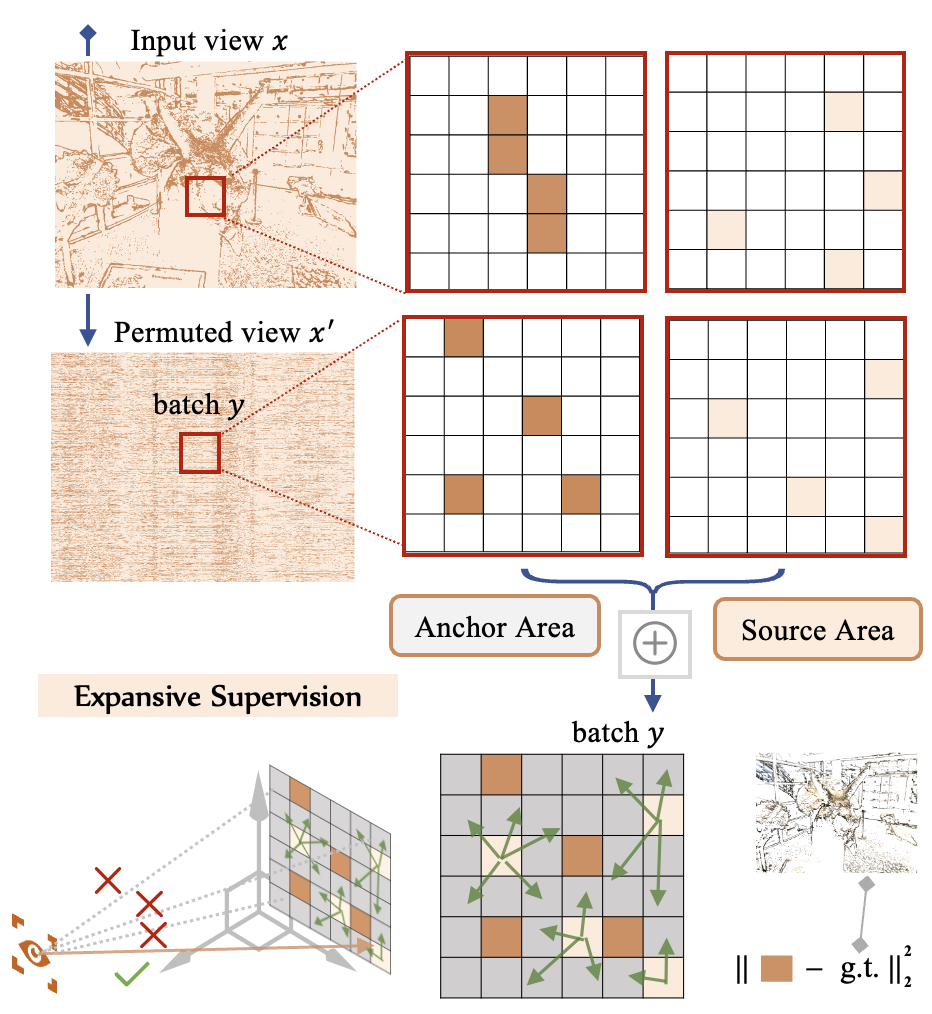}}
    \caption{\textbf{Pipeline of expansive supervision.}
    The mechanism of expansive supervision is to exclusively render the crucial pixels, which consist of the pre-computed anchor area and sampled source areas, to estimate the loss. This estimation is accomplished through the expansive strategy described in Section \ref{sec:supervision}. 
    }
    \label{fig:pipeline}
\end{figure}

\subsection{Expansive Supervision}
\label{sec:supervision}
The objective of \textit{Expansive Supervision} is to utilize an important subset of supervision signals to guide radiance field training while maintaining rendering quality and achieving substantial time and memory savings. Our approach is motivated by the observation that error distribution strongly correlates with image content: regions with high-frequency details exhibit larger training errors and require more attention.

This observation is validated through a preliminary study, as illustrated in Fig.~\ref{fig:observation}. During standard training, the error distribution exhibits a clear long-tail phenomenon, where approximately 99.4\% of the data points with the smallest errors contribute only 10\% of the overall loss. Moreover, higher error values predominantly concentrate in high-frequency image regions, particularly edges and texture-rich areas, as demonstrated in column 3 of Fig.~\ref{fig:observation}. These findings suggest that the global error distribution can be effectively estimated from partial renderings, enabling expansive loss calculation.

The pipeline of Expansive Supervision is demonstrated in Fig.~\ref{fig:pipeline}. 
The selected pixel set $R^{\prime}$ comprises two distinct areas within the input view $I$:

\textbf{1) Anchor area} $A \subset I$. The anchor area $A$ represents regions within $I$ that exhibit larger error patterns. It is computed using the anchor extractor function $\mathcal{F}{A}(\cdot)$, which is detailed in Sec.~\ref{sec:extractor}. Formally, $A = \mathcal{F}{A}(I, \xi_a)$, where $\xi_a$ controls the anchor area size such that $|A| = \xi_{a} |I|$.

\textbf{2) Source area} $S \subset { I \backslash A }$. The source area $S$ consists of sampled points from the remaining regions after excluding the anchor set. It is formulated as $S \sim \mathcal{U}_{\xi_s|I|}({ I \backslash A })$, where $\mathcal{U}(\cdot)$ denotes uniform sampling and $\xi_s$ controls the source area size such that $|S| = \xi_{s} |I|$. The source area is regenerated for each iteration, and its errors are expanded to the remaining regions to achieve comprehensive loss estimation.

We define $B^*_t$ as the batch data after random permutation at iteration $t$, with $A^*_t$ and $S^*_t$ representing its corresponding anchor and source areas. The global estimated error $\hat{L_t}$ at iteration $t$ can be represented as:
\begin{equation}
    \label{eq:error}
    \begin{aligned}
    \hat{L_t} = & \underbrace{\sum_{r_a \in A^*_i }||\hat{C}(r_a) -C(r_a)||^{2}_{2}}_{\text{anchor supervision}} + \\ 
    & \underbrace{[\gamma_{s/a}+ \frac{t}{T}(1-\gamma_{s/a})]}_{\text{expansive mechanism }} \underbrace{\sum_{r_s \in S^*_i }||\hat{C}(r_s) -C(r_s)||^{2}_{2}}_{\text{source supervision}},
    \end{aligned}
\end{equation}
where $C (\cdot)$ and $\hat{C}(\cdot)$ denote ground truth and predicted RGB colors from given rays.
The expansive factor $\gamma{s/a}$ is computed as ${(1-\xi_a)}/{\xi_a}$, and $T$ denotes the total number of iterations.
At each iteration's end, the parameter $\Theta$ is updated via $\Theta:=\Theta-\eta \nabla \hat{L}_t$. Compared to standard full supervision, Expansive Supervision renders only a subset of rays ${R}^{\prime} = {A \cup S} \subset {R}$ to guide the model learning process. This selective rendering theoretically reduces computational resources (time and memory) proportional to $\xi$, where $\xi = \xi_a + \xi_s = |R^{\prime}| / |R|$.

\subsection{Anchor Area Extractor}
\label{sec:extractor}
Here we detail the design of the anchor area extractor $\mathcal{F}_{A}(\cdot)$. 
Given that Expansive Supervision aims to accelerate training, we employ the lightweight edge detector~\cite{canny} as our basic extractor for frequency prior. 
To maintain consistent anchor area intensity of $\xi_a|I|$, we implement a progressive threshold adjustment mechanism for the edge detector.
At iteration $i$, the threshold $T_i$ is updated according to:
\begin{equation}
T_i = 1 + \mu(\mathcal{E}(T_{i-1}) - \xi_a|I|)
\end{equation}
where $\mathcal{E}(T_{i-1})$ represents the sum of the detector output map with threshold $T_{i-1}$, and $\mu$ denotes the step rate. The iteration continues until the condition $0.8 \leqslant \frac{\mathcal{E}(T_{i})}{\xi_a|I|} \leqslant 1.2$ is satisfied.

%% file: content/4-experiments.tex
\begin{table*}[t]
    \centering
    \caption{\textbf{Quantitative comparison of different strategies on Synthetic-NeRF Datasets.}
    }
    \label{table:supervision}
    \begin{tabular}{l|ccc|cc|cc|ccc|cc}
    \toprule
     & \multicolumn{3}{c|}{Iteration: 5k} & \multicolumn{2}{c|}{Iteration: 10k} &  \multicolumn{2}{c|}{Iteration: 15k} & \multicolumn{3}{c|}{Iteration: 30k} & \multicolumn{2}{c}{Consumption}\\ 
    
    Strategies & PSNR   & SSIM  & L(V) & PSNR   & SSIM  & PSNR   & SSIM   &  PSNR  & SSIM  & L(V)  & Time & Memo.\\
    
    & (dB)~$\uparrow$ & $\uparrow$ & $\downarrow$ & (dB)~$\uparrow$ & $\uparrow$ & (dB)~$\uparrow$ & $\uparrow$  & (dB)~$\uparrow$ & $\uparrow$ & $\downarrow$   & (sec)~$\downarrow$ & (GB)~$\downarrow$ \\ 
    
    \midrule \midrule
    
    Standard &  29.30 & 0.933 & 9.52e-2 &  30.78 & 0.946 
    & 31.78 &\cellcolor{yellow}0.954  & 32.91 & \cellcolor{brown}0.962  & \cellcolor{brown}4.97e-2 & 587.25 & 21.47\\  
    \midrule
    EGRA~\cite{egra} & 29.33 & 0.933 & 9.48e-2 & 30.81 & 0.947  & 31.74 & \cellcolor{yellow}0.954 & 32.85 & \cellcolor{yellow}0.961  & \cellcolor{yellow}5.10e-2 &  1771.85 & 17.10\\ 
    Soft~\cite{soft} & 29.44 & \cellcolor{yellow}0.938 & \cellcolor{yellow}9.35e-2 & 30.88 & \cellcolor{brown}0.950 & 31.61 & \cellcolor{brown}0.955 & 32.34 & 0.959  & 6.05e-2 & 585.32 & 16.29\\
    E.S.~(ours)&  \cellcolor{brown}30.36 & \cellcolor{brown}0.939 & \cellcolor{brown}9.20e-3 & \cellcolor{brown}31.77 & \cellcolor{yellow}0.949 &  \cellcolor{brown}32.42& \cellcolor{yellow}0.954& \cellcolor{brown}33.02 & \cellcolor{yellow}0.961  & 6.12e-2 & 578.54  & 18.33 \\
    \midrule
    E.S.~($\beta = 0.9$) & \cellcolor{yellow}30.21 & \cellcolor{yellow}0.938 & 9.37e-2 & \cellcolor{yellow}31.65 & 0.948  & \cellcolor{yellow}32.36 & 0.953 & \cellcolor{yellow}32.96 & 0.955  & 6.04e-2  & 576.17 & 16.78 \\
    E.S.~($\beta = 0.7$) & 29.92 & 0.934 & 9.85e-2 & 31.42 & 0.946  & 32.16 & 0.951 & 32.87 & 0.954  & 6.25e-2 & 529.47  & 12.57 \\
    E.S.~($\beta = 0.5$)\textsuperscript{$\dagger$}  & 29.41 & 0.929 & 1.06e-1 & 30.95 & 0.941 & 31.77 & 0.948 & 32.64 & 0.952  & 6.56e-2 & 491.32 & 10.37 \\ 
    E.S.~($\beta = 0.3$)\textsuperscript{$\dagger$} & 28.73 & 0.922 & 1.17e-1 & 30.30 & 0.935  & 31.24 & 0.943 & 32.31 & 0.950  & 6.93e-2 & 438.91 & 6.29\\ 
    E.S.~($\beta = 0.1$) & 26.74 & 0.898 & 1.52e-2 &  28.29 & 0.914  & 29.37 & 0.925& 30.86 & 0.939  & 8.51e-2 & 388.46 & 2.56\\ 
    \bottomrule
    \multicolumn{13}{l}{\scriptsize{$\dagger$ denotes the settings strike a balance between quality and efficiency.
    \colorbox{brown}{The best result};\colorbox{yellow}{The second best result}. The following tables use the same notation.}}
    \end{tabular}
\end{table*}

\section{Experiments}

\subsection{Implementation Details}

\noindent \textbf{Backbones \& Hyper-parameters.}
To demonstrate the compatibility of our method with state-of-the-art NeRF acceleration frameworks, we employ TensoRF~\cite{tensorf} as the backbone model for all experiments. The key hyper-parameters are set as follows: $\xi_a = \xi_s = 0.25$, $T=30,000$, and $\mu = 15$. 
The ray filtering operation is disabled to ensure compatibility with various strategies.
We introduce a quality-computation trade-off parameter $\beta \in [0,1]$ that modulates the supervision intensity through $\xi_a := \beta \times \xi_a$ and $\xi_s := \beta \times \xi_s$, where higher $\beta$ values yield better reconstruction quality at the cost of increased computational resources.
All other parameters follow the default TensoRF configuration.

\noindent \textbf{Datasets \& Metrics.}
We evaluate our method using both the object-centric Synthetic-NeRF~\cite{nerf} and real-world forward-facing LLFF~\cite{llff} datasets. For quantitative assessment, we employ PSNR, SSIM, and LPIPS~\cite{lpips} metrics, where L(V) denotes the  VGG~\cite{vgg} versions of LPIPS.

\noindent \textbf{Experimental Settings.}
All experiments were conducted using PyTorch~\cite{torch} on a system equipped with four NVIDIA RTX 3090 GPUs (24.58GB VRAM).

\begin{table}[tbp]
    \caption{
    {\textbf{comparison of different strategies on LLFF Datasets.}}
    }
    \label{table:llff_compare}
    \centering
    \begin{tabular}{l|c|c|ccc}
        \toprule
         & \multicolumn{1}{c|}{5k} & \multicolumn{1}{c|}{10k} & \multicolumn{3}{c}{20k} \\ 
        & PSNR$\uparrow$ & PSNR$\uparrow$ & PSNR$\uparrow$ & SSIM$\uparrow$ & L(V)$\downarrow$  \\
        \midrule
        Standard  & 23.86 & 25.43 & 26.10 & \cellcolor{yellow}0.813  & 0.241\\
        \midrule
        EGRA~\cite{egra}  & 23.83 &25.30 & 26.08 & 0.811 & 0.237 \\
        Soft~\cite{soft}  & \cellcolor{yellow}23.94 & \cellcolor{yellow}25.47 &  25.96 & \cellcolor{brown}0.819 & \cellcolor{brown}0.227 \\
        E.S.~($\beta=1.0$) &  \cellcolor{brown}24.11 & \cellcolor{brown}25.60 & \cellcolor{brown}26.13 & \cellcolor{yellow}0.813 & \cellcolor{yellow}0.236 \\
        \midrule
        E.S.~($\beta = 0.7$)  & 23.83  & 25.41 &  \cellcolor{yellow}26.11  &0.812   & 0.243  \\
        E.S.~($\beta = 0.5$)  & 23.57  & 25.21 &  26.01  &0.811    & 0.250  \\
        \bottomrule
    \end{tabular}
\end{table}

\begin{figure}[!t]
    \centerline{\includegraphics[width=\columnwidth]{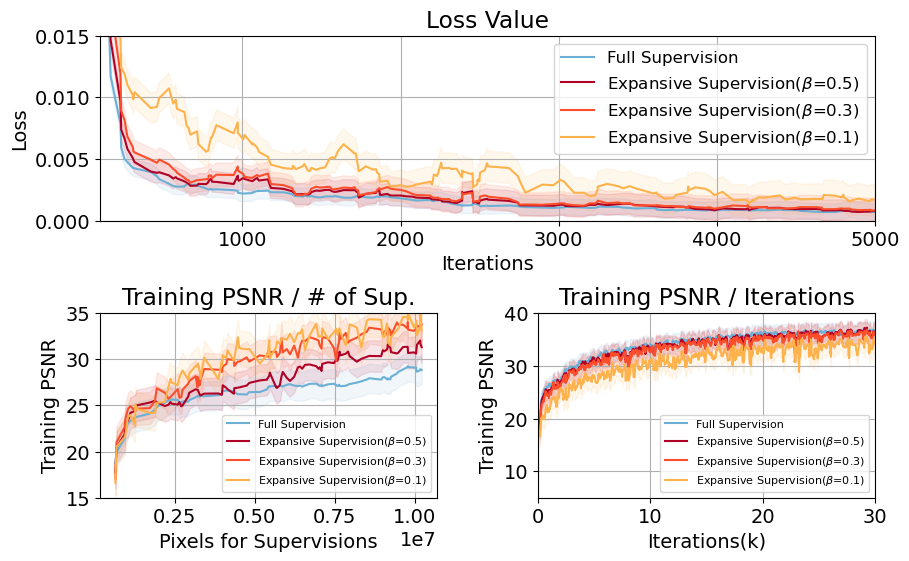}}
    \caption{\textbf{Convergence performance of expansive supervision.} Our method achieves precise error estimation comparable to full supervision and exhibits faster convergence as the number of supervised pixels increases.}
    \label{fig:convergence}
\end{figure}

\begin{figure*}[!t]
    \centerline{\includegraphics[width=\textwidth ]{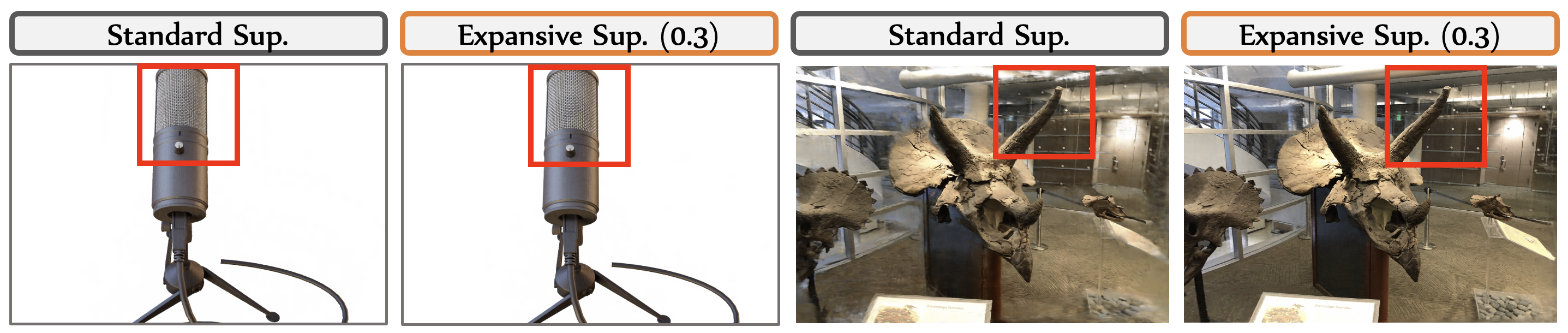}}
    \caption{\textbf{Visual quality comparison with standard supervision}. 
    Under the same constrained computational resources, expansive supervision demonstrates higher quality reconstruction compared to standard supervision.
    }
    \label{fig:visual}
\end{figure*}

\subsection{Comparison with State-of-the-arts Strategies}

\noindent \textbf{Comparing Baselines.}
We compare our method with standard training (random batch sampling) and recent sampling-based NeRF acceleration methods: EGRA~\cite{egra} and Soft Mining~\cite{soft}. For EGRA, we re-implement the algorithm following the original paper's specifications, as their official implementation is not available. For Soft Mining, we directly adapt their official implementation without modifications, maintaining their recommended soft coefficients ($\alpha=0.6$ for Synthetic-NeRF and $\alpha=0.8$ for LLFF datasets).
Time measurements exclude I/O operations and preprocessing costs. Memory consumption during training is recorded and reported as Memo.

\noindent \textbf{Quantitative Results.} The quantitative comparison results on Synthetic-NeRF and LLFF datasets are presented in Tab.\ref{table:supervision} and Tab.\ref{table:llff_compare}, respectively. Compared to existing sampling-based acceleration methods, Expansive Supervision (E.S.) achieves superior reconstruction quality under comparable time and memory constraints, with particularly notable improvements during early training stages. This performance advantage is consistently demonstrated across both synthetic and real scenes. Furthermore, our ablation study with varying $\beta$ values shows that with $\beta=0.5$, our method achieves comparable reconstruction quality while reducing training time by 84\% and memory consumption by 48\% relative to standard training.

\noindent \textbf{Convergence \& Visualization Results.}
The convergence performance of our method is visualized in Fig.~\ref{fig:convergence}, demonstrating the effectiveness of Expansive Supervision. Our approach with $\beta=0.3$ and $\beta=0.5$ achieves error estimation accuracy comparable to full supervision, as evidenced in the upper and lower right sub-figures. The convergence rate accelerates as the number of supervised pixels increases.
The visual quality comparison is presented in Fig.~\ref{fig:visual}. We evaluate our method against full supervision under varying computational constraints, simulated by adjusting batch sizes while maintaining consistent iteration counts. Our method, with its emphasis on high-frequency regions, achieves superior detail preservation compared to standard supervision under similar computational budgets.

\subsection{Analysis of Resource Savings}
\label{sec:tradeoff}
\begin{figure}[tbp]
    \centerline{\includegraphics[width=\columnwidth]{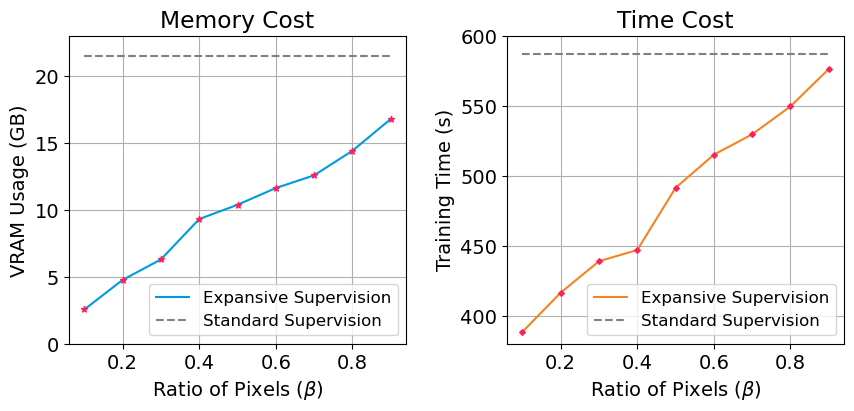}}
    \caption{\textbf{Memory and training time cost of expansive supervision.}
    Both memory and time consumption demonstrate linear scaling with $\beta$
    }
    \label{fig:visual cost}
\end{figure}

\noindent \textbf{Measurements.}
To validate the theoretical resource savings outlined in Sec.~\ref{sec:formulation}, we conducted systematic measurements of practical memory and time consumption under Expansive Supervision. The experiments were performed in isolation on a dedicated server, focusing exclusively on training operations (forward passes, loss computation, backward passes, and volume rendering), excluding I/O and pre-processing overhead.
To ensure measurement precision, all extraneous processes were terminated, and experiments were conducted sequentially. 
We evaluated computational costs across ten $\beta$ values ranging from 0.1 to 1.0, with results visualized in Fig.~\ref{fig:visual cost}. Both memory and time consumption demonstrate linear scaling with $\beta$, confirming its effectiveness as a control parameter for balancing reconstruction quality against computational resources.

\begin{figure}[!t]
    \centerline{\includegraphics[width=\columnwidth]{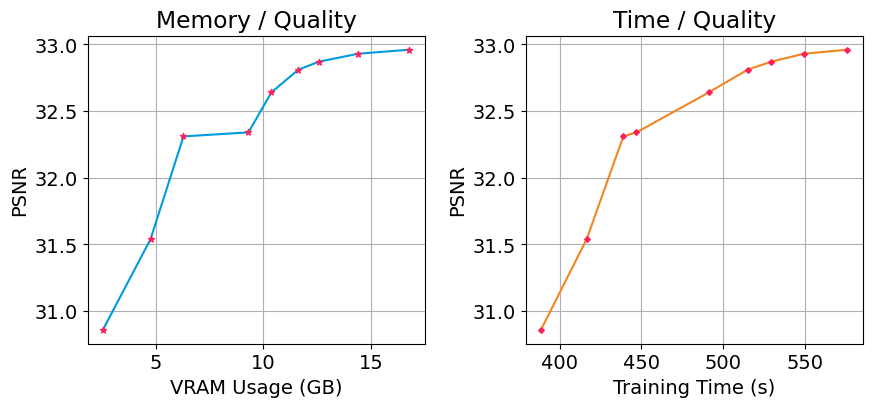}}
    \caption{\textbf{Memory and training time cost of expansive supervision.}
    As the supervised pixel ratio increases, the margin of resource savings decreases.}
    \label{fig:tradeoff}
\end{figure}

\noindent \textbf{Time-Quality Trade-off.}
We analyze the relationship between resource utilization and model performance across various configurations. The resource-quality curves illustrated in Fig.~\ref{fig:tradeoff} demonstrate how $\beta$ modulates both time and memory consumption.
Our analysis reveals an optimal reduction in supervised pixels at approximately $\beta=0.3$, where $\frac{\text{d}C(\text{Time})}{\text{d}\beta}$ and $\frac{\text{d}C(\text{Memo.})}{\text{d}\beta}$ reach their maximal values. Beyond this point, the marginal resource savings diminish significantly. While lower supervision ratios ($\beta=0.1$) lead to substantial resource savings, they introduce noticeable artifacts incompatible with our mechanism. Thus, $\beta=0.3$ emerges as the optimal trade-off between rendering quality and computational efficiency.

\subsection{Extension to Implicit Neural Representations for Images}
\noindent \textbf{Settings.}
To further validate the effectiveness and generalizability of our method, we extend it to Implicit Neural Representations (INR) for images. Image fitting can be formulated as $F(\Theta): (x,y) \mapsto (r,g,b)$. We employ SIREN~\cite{siren} with a $3\times256$ network architecture as the backbone, evaluated on the processed DIV2K dataset~\cite{div2k, sym-trans}. Training proceeds for 5,000 iterations with a batch size of $0.5\times h \times w$, where $h$ and $w$ denote the image height and width, respectively.

\noindent \textbf{Results.}
The comparative results, averaged across the dataset, are presented in Tab.~\ref{table:inr}. Our method outperforms standard training~(uniform sampling), EGRA, and Soft Mining, with particularly notable improvements in later training stages. These results demonstrate that Expansive Supervision generalizes beyond NeRF applications to other neural representation modalities.

\subsection{Ablation Studies}
\noindent \textbf{Settings.}
To validate the effectiveness of each component within our proposed Expansive Supervision mechanism, we conducted comprehensive ablation experiments. Using Expansive Supervision ($\beta=1.0$) as the baseline, we first examined the impact of different sampling areas: \textit{w/o anchor area} samples only from non-important regions, while \textit{w/o source area} samples exclusively from fixed anchor areas. Furthermore, we evaluated each component in Eq.~\ref{eq:error}: anchor supervision, source supervision, and the expansive mechanism.

\noindent \textbf{Results.}
The results presented in Tab.~\ref{table:ablations} demonstrate that each component contributes distinctly to the overall performance. Removing either anchor or source areas effectively reduces our method to uniform sampling in different spaces. The \textit{w/o source area}, which samples only from anchor areas, shows severe performance degradation due to its limited sampling space.
In the analysis of Eq.~\ref{eq:error} components, removing source supervision (\textit{w/o source sup.}) introduces significant training bias by eliminating uncertainty supervision. The \textit{w/o expansive} configuration, which merely supervises two areas independently, reduces Eq.~\ref{eq:error} to approximate squared error, resulting in suboptimal reconstruction quality.

\begin{table}[tbp]
    \caption{
    {\textbf{Comparison on 2D image Fitting with INRs.}}
    }
    \label{table:inr}
    \centering
    \begin{tabular}{l|c|c|ccc}
        \toprule
         & \multicolumn{1}{c|}{1k} & \multicolumn{1}{c|}{2k} & \multicolumn{3}{c}{5k}\\ 
        & PSNR$\uparrow$ & PSNR$\uparrow$ & PSNR$\uparrow$ & SSIM$\uparrow$ & L(V)$\downarrow$  \\
        \midrule
        Standard  & 30.35 & 33.33 & 36.11 & \cellcolor{yellow}0.956 & 0.026 \\
        \midrule
        EGRA~\cite{egra}  & 30.37 & 33.40 & \cellcolor{yellow}36.22 & \cellcolor{brown}0.957 & \cellcolor{yellow}0.025    \\
        Soft~\cite{soft}  & \cellcolor{brown}31.49 & \cellcolor{yellow}33.52 & 35.31 &  0.948 & 0.045\\
        E.S.~($\beta=1.0$)&  \cellcolor{yellow}30.95 & \cellcolor{brown}33.70 & \cellcolor{brown}36.28 & 0.955 & \cellcolor{brown}0.023 \\
        \bottomrule
    \end{tabular}
\end{table}

\begin{table}[tbp]
    \caption{
    {\textbf{Ablation Study.}}
    }
    \label{table:ablations}
    \centering
    \begin{tabular}{l|c|cccc}
        \toprule
        & \multicolumn{1}{c|}{5k} & \multicolumn{4}{c}{30k} \\ 
        Settings & PSNR$\uparrow$  & PSNR$\uparrow$ & SSIM$\uparrow$ & L(A)$\downarrow$ & L(V)$\downarrow$  \\
        \midrule
        E.S.~($\beta=1.0$)  & \cellcolor{brown}30.36 & \cellcolor{brown}33.02  & \cellcolor{brown}0.961   & \cellcolor{brown}3.69e-2 & \cellcolor{brown}6.12e-2 \\
        \midrule
        w/o anchor area   &27.29 &29.68   &0.946    &4.53e-2 & 6.41e-2\\
        w/o source area & 17.36  &18.34  & 0.451   &6.14e-1 & 5.24e-1\\
        \midrule
        w/o anchor sup.  &26.59  &29.10  &0.939     &5.45e-2 & 7.38e-2 \\
        w/o source sup.   &19.34  &20.21  &0.586    &5.28e-1 &4.59e-1 \\
        w/o expansive  & 29.98 & 32.69 &  0.949 & 4.40e-2 & 8.56e-2 \\ 
       
        \bottomrule
    \end{tabular}
\end{table}

%% file: content/5-conclusion.tex
\section{Conclusion}
In this paper, we introduce an expansive supervision mechanism that enhances the efficiency of neural radiance field training. Our approach leverages the observation that training error distributions exhibit strong correlations with image content, following a long-tail distribution pattern. By strategically selecting ray subsets for rendering and utilizing image context for expansive error estimation, our method achieves significant computational efficiency while maintaining reconstruction quality.
Experimental results demonstrate that our method outperforms existing sampling-based acceleration approaches under equivalent computational constraints. Moreover, when integrated with current acceleration frameworks, our approach achieves substantial memory savings with minimal implementation effort.